\documentclass[11pt]{article}
\usepackage{amsmath,amssymb}
\usepackage{graphicx}
\usepackage{booktabs}
\usepackage{url}
\usepackage[margin=2.5cm]{geometry}
\usepackage{pgfplots}
\pgfplotsset{compat=1.17}
\usepgfplotslibrary{fillbetween} %
\usepgfplotslibrary{groupplots}  %
\pgfplotsset{discard if not/.style 2 args={x filter/.append code={%
  \edef\tempa{\thisrow{#1}}\edef\tempb{#2}%
  \ifx\tempa\tempb\else\fi}}}

\newcommand{\FSet}{Y}                 %
\newcommand{\Dim}{D}                  %
\newcommand{\Sub}[1]{X_{#1}}          %
\newcommand{\Crit}{J}                 %
\newcommand{\ADD}[1]{\mathrm{ADD}_{#1}}
\newcommand{\RMV}[1]{\mathrm{RMV}_{#1}}
\newcommand{\sADD}[2]{\mathrm{sADD}^{(#1)}_{#2}}
\newcommand{\muis}[1]{\mu_{#1}}       %
\newcommand{\muisnot}[1]{\bar{\mu}_{#1}}
\newcommand{\score}[1]{s(#1)}         %
\newcommand{\budget}{y}               %

\title{Stochastic Sequential Search in Very-High-Dimensional Feature Selection}
\author{Petr Somol \quad Ji\v{r}\'{\i} Grim\\[4pt]
  {\normalsize Institute of Information Theory and Automation,
   The Czech Academy of Sciences}\\
  {\normalsize Prague, Czech Republic}}
\date{}

\begin{document}
\maketitle

\begin{abstract}
Sequential subset search --- forward selection with floating backtracking and
its descendants --- remains the quality reference in feature selection, but
every member of the family sweeps the full pool of remaining candidate
features at each step, which
excludes it from very-high-dimensional problems. There, individual-feature
ranking methods are the only practical option, and they are sub-optimal by
design: feature interplay is modeled weakly or not at all. We introduce a
budgeted sampled step operator pair that replaces the full sweeps of
sequential search by a fixed number of candidate evaluations per step.
Candidates are drawn by temperature-controlled softmax sampling from
dependency-aware per-feature statistics learned online from every criterion
evaluation the search itself performs, guarded by a uniform exploration floor
that keeps every feature reachable at every step; per-step cost becomes
independent of dimensionality. Substituting the operators turns any sequential
method into its stochastic counterpart, defining the Stochastic Sequential
Search (SSS) family; we study the stochastic counterpart of floating search
(sSFFS). On the 500-dimensional madelon benchmark, sSFFS retains at
least 97\% of the full-SFFS criterion value at every subset size up to 165
while spending about one quarter of its criterion evaluations; uniform
sampling at the same budget collapses on madelon's synergistic features, and
freezing the statistics after warm-up loses significantly --- the online
adaptation carries the quality. On the 5{,}000-dimensional gisette,
sSFFS sustains criterion values near $0.99$ at every subset size
up to 320 --- far beyond full-SFFS reach --- exceeding the saturated level
of DAF and BIF ranking on the search objective by $0.05$--$0.10$ at
matched evaluation budgets; holdout validation of the same subsets shows
that at 500 training samples this optimization power outruns what the
wrapper criterion can support, an honest demonstration that beyond the
sequential frontier the binding constraint becomes the criterion, not the
search. On the $10{,}105$-dimensional reuters corpus, under a multinomial
filter criterion that deserves trust, the advantage transfers to holdout
intact: sSFFS dominates BIF and DAF on the search objective
\emph{and} on holdout accuracy at every subset size (e.g.\ $0.867$
vs.\ $0.829$/$0.837$ at subset size 100), with the entire run costing about two
minutes of single-core evaluation work. The method is implemented in the
authors' Feature Selection Toolbox; a verified standalone implementation
and the configurations of all experiments accompany the paper.
\end{abstract}

\section{Introduction}
\label{sec:intro}

Feature selection under a criterion that reflects feature behavior \emph{in
context} --- most prominently the wrapper setting, where candidate subsets are
scored by the accuracy of a classifier trained on them \cite{kohavi97} ---
faces a hard computational frontier. Optimal search is confined to a few tens
of dimensions; sequential (hill-climbing) subset search extends to the
hundreds; beyond that, only individual-feature ranking remains practical
\cite{somol11daf}. The methods on the far side of this frontier are
sub-optimal \emph{by design}: Best Individual Features (BIF) ranking ignores
feature interplay entirely --- praised for stability \cite{kuncheva07}, yet it
``fails completely'' on data with synergistic features \cite{somol11daf} ---
and even dependency-aware ranking (DAF) \cite{somol11daf}, which recovers
interaction information from randomly probed subsets, produces a ranking, not
a constructed subset: no set-level decision is ever made. Meanwhile problems
with thousands to tens of thousands of features --- text categorization,
microarray analysis --- have long been routine.

The frontier has barely moved in fifteen years, because it is set by the
search's own economics rather than by hardware. In 2011, full floating search
(SFFS \cite{pudil94}) needed 16 minutes per trial on the 57-dimensional
spambase but could not finish a single trial on 500-dimensional madelon in
five days on a 16-CPU system \cite{somol11daf}. Our 2026 single-thread
reference completes the experimental madelon size range (all subset sizes up
to 165; the criterion optimum lies at subset size $d=21$) in roughly five
hours ---
hardware bought one order of magnitude. But the cost of every sequential method still contains
the per-step full sweep: with $\Dim$ features in total, $O(\Dim-d)$ criterion
evaluations per forward step at working-subset size $d$, $O(d)$ per backward
step, with per-evaluation cost itself growing with subset size. On gisette (5{,}000-dim) the measured 30-minute frontier of SFFS lies
near subset size 34, subset size 300 costs on the order of a day, and running
to completion is out of the question; at $\Dim\gtrsim 10^4$ (text domains) no
amount of foreseeable hardware rescues the sweep. The frontier moves only if
the per-step economics change.

This paper changes them. We replace the exhaustive sweep of the sequential
step by a \emph{budgeted sampled step}: exactly $\budget$ candidates are
evaluated per step, drawn without replacement from a proposal distribution
that the search \emph{learns online} --- dependency-aware per-feature
statistics in the spirit of DAF \cite{somol11daf}, accumulated at zero
additional cost from the criterion evaluations the search performs anyway,
z-standardized within each step's batch and aged with an exponential
forgetting horizon. A temperature-controlled softmax concentrates the budget
on currently-promising features; a uniform exploration floor
($\rho_u$ of each batch, drawn from the least-evaluated features) keeps every
feature reachable at every step, so early mis-rankings are self-correcting
rather than permanent. The operator pair is direction-symmetric
($\mathrm{sADD}$~/~$\mathrm{sRMV}$, Section~\ref{sec:method}), and
substituting it for the classical
$\ADD{}$/$\RMV{}$ operators turns \emph{any} member of the sequential family
--- sequential forward/backward selection (SFS/SBS), their floating
counterparts (SFFS/SBFS), the oscillating variants --- into a budgeted method
whose per-step cost is independent of $\Dim$. We call the resulting family
\emph{Stochastic Sequential Search} (SSS)\footnote{Not to be confused with
the game-tree search algorithm SSS* of Stockman \cite{stockman79}; the
acronym here names a feature-selection family.} and name its members by prefixing
the classical acronym with a lowercase s for stochastic --- the same s that
marks the substituted operators $\mathrm{sADD}$/$\mathrm{sRMV}$: sSFS,
sSBS, sSFFS, sSBFS. This
paper studies \textbf{sSFFS}, the stochastic counterpart of floating search.
The search remains anytime and
$d$-optimizing: it reports the best subset found at every size.

A fixed uniform per-step sample is known to approximately preserve greedy
quality for monotone submodular objectives \cite{mirzasoleiman15lazier};
the in-context criteria of feature selection --- wrapper accuracy chief
among them --- are not submodular, and precisely there uniform sampling
fails --- synergistic features that look worthless individually are found only
by luck, and the failure is empirically catastrophic, not graceful
(Section~\ref{sec:results}: on madelon, uniform budget spending collapses to
chance-level validated accuracy on some seeds). The design response is to make
the sampler \emph{informed}: the online statistics recover exactly the
interaction signal that uniform sampling misses, while sampling (as opposed to
deterministic truncation of a static ranking, the choice of pool-restriction
prior art \cite{gutlein09lfs,somol10hybrids}) preserves reachability of every
feature. Our ablation matrix is built so that each of these design choices is
falsifiable in isolation: the uniform sampler \emph{is} stochastic-greedy
transferred to sequential FS, the top-$k$ sampler \emph{is} the hybridization
principle with statistics-as-filter, and the frozen-statistics variant
\emph{is} the static-ranking principle of LFS --- each is a named prior-art
principle embedded as a switch in the same implementation.

We deliberately do not position budgeted search against full sequential search
--- where the full method is tractable it should be preferred, and comparisons
are only computable there. Full-SFFS comparisons on madelon serve as
\emph{retention evidence}: sSFFS with default knobs retains
$\ge 97\%$ of the full-SFFS criterion value at every subset size up to 165 on
the shared data split of the full-search reference, at roughly one quarter of
the evaluations (with both
directions budgeted). The value proposition lies beyond the frontier, where
the only competitors are ranking methods: there the relevant comparison is
against DAF and BIF at \emph{matched evaluation budgets}, and the question the
experiments answer is how much budget a sequential method needs before it
overtakes ranking. On gisette the answer on the optimization axis is: the
smallest we tried --- sSFFS exceeds the ranking methods' saturated
criterion level at every budget in the ladder, while holdout validation of
the same subsets locates the regime's remaining constraint in the wrapper
criterion itself rather than in the search
(Section~\ref{sec:results}). On the $10{,}105$-dimensional reuters corpus,
finally, where the criterion is a trustworthy multinomial filter rather
than a cross-validation proxy, the budgeted search dominates BIF and DAF
on \emph{both} axes at every subset size --- the optimization advantage
transfers to holdout intact, and the full frontier run costs about two
minutes of single-core evaluation work (Section~\ref{sec:results}).

The contributions are:
\begin{enumerate}
\item a direction-symmetric budgeted sampled step operator pair
  ($\sADD{\budget_f}{c}$, $\mathrm{sRMV}^{(\budget_b)}_c$; $\budget_f$,
  $\budget_b$ the forward/backward per-step budgets, $c$ the classical step
  tuple size) with per-step cost independent of $\Dim$, turning every
  classical sequential method into a budgeted variant by operator
  substitution (Section~\ref{sec:method});
\item an online dependency-aware statistic --- batch-standardized criterion
  contrasts with exponential forgetting, updated from every evaluation the
  search itself performs --- requiring no second criterion and no additional
  evaluations (Section~\ref{sec:method});
\item a proposal policy (softmax with robust automatic temperature, uniform
  exploration floor) that avoids both the synergy-blindness of uniform
  sampling and the permanence of deterministic truncation;
\item an experimental map of the quality-vs-compute trade-off of budgeted
  floating search with ranking-method anchors, plus ablations isolating each
  ingredient against its prior-art principle (Section~\ref{sec:results});
\item command-line-level reproducibility of every reported experiment
  through a verified standalone open-source implementation and the
  configuration files and command lines accompanying the paper.
\end{enumerate}

Section~\ref{sec:related} places the method in the sequential, pool-restriction,
submodular, and bandit lineages; Section~\ref{sec:method} defines the operators
and statistics; Sections~\ref{sec:setup}--\ref{sec:results} report the
experimental campaign; Section~\ref{sec:discussion} discusses parameter
sensitivity and limitations.

\section{Preliminaries and Related Work}
\label{sec:related}
Let $\FSet=\{f_1,\dots,f_\Dim\}$ denote the full feature set and
$\Sub{d}\subset\FSet$ a working subset of cardinality $d$, evaluated by a
criterion $\Crit(\cdot)$. Sequential methods are compositions of two
operators \cite{somol11kybernetika}: $\ADD{c}$ joins to the working subset
the conditionally most significant feature $c$-tuple, $\RMV{c}$ discards the
conditionally least significant one,
\begin{align}
\ADD{c}(\Sub{d}) &= \Sub{d}\cup\mathcal{T}^{+}, &
\mathcal{T}^{+} &= \arg\max_{\substack{\mathcal{T}\subseteq\FSet\setminus\Sub{d},\;|\mathcal{T}|=c}}
\Crit(\Sub{d}\cup\mathcal{T}),\notag\\
\RMV{c}(\Sub{d}) &= \Sub{d}\setminus\mathcal{T}^{-}, &
\mathcal{T}^{-} &= \arg\max_{\substack{\mathcal{T}\subseteq\Sub{d},\;|\mathcal{T}|=c}}
\Crit(\Sub{d}\setminus\mathcal{T}),\label{eq:addrmv}
\end{align}
with repeated application written $\ADD{c}^{\delta}$ and $\RMV{c}^{\delta}$;
$c{=}1$ throughout unless stated otherwise. SFS is the pure composition
$\Sub{t}=\ADD{c}^{\delta}(\emptyset)$; SFFS \cite{pudil94} interleaves the
two conditionally. For a target size $t$ and an optional frontier
restriction $\Delta\in[0,\Dim-t]$:
\begin{enumerate}
\item start with $\Sub{c}=\ADD{c}(\emptyset)$, $d=c$;
\item forward: $\Sub{d+c}=\ADD{c}(\Sub{d})$, $d\leftarrow d+c$;
\item retreat: repeat $\Sub{d-c}=\RMV{c}(\Sub{d})$, $d\leftarrow d-c$, but
      only while each removal improves the best value recorded so far at the
      lower cardinality;
\item if $d<t+\Delta$ go to step~2; else report the best recorded $\Sub{t}$.
\end{enumerate}
The working subset thus advances and retreats: it grows greedily, and after
every
growth step it sheds whatever earlier choices the newest additions have made
redundant; SBFS mirrors the scheme from the full set downward, with the
operator roles exchanged. This conditional backtracking is what frees floating search from
the nesting problem of plain SFS \cite{devijver82}, at the price of the
extra $\RMV{}$ sweeps; $\Delta$ bounds how far past $t$ the frontier may
grow, the classic algorithm running unrestricted ($\Delta=\Dim-t$).

Oscillating search (OS) \cite{somol00} complements the floating scheme when
the target cardinality $t$ is fixed in advance. It maintains a \emph{pivot}
--- the best known subset of size $t$, initialized by any means available
(typically the prefix of an individual-feature ranking, or another search's
result) --- and repeatedly \emph{swings} around it: a down-swing applies
$\RMV{c}$ $o$ times and then $\ADD{c}$ $o$ times, probing whether $o$
exchanged features improve the pivot from below; an up-swing mirrors the
excursion upward through intermediate sizes $t{+}c,\dots,t{+}o\,c$. Only
when a completed swing has returned to size $t$ is its result compared with
the pivot: an improvement makes it the new pivot and resets the swing depth
to $o{=}1$, while a failed down/up pair deepens the next swings,
$o\leftarrow o{+}1$, until $o$ would exceed a user-set limit $o_{\max}$ and
the search stops (the dynamic variant DOS \cite{somol08dos} additionally
lets the pivot cardinality itself drift, optimizing $t$ alongside the
subset). Where floating search constructs solutions at every cardinality on
its way to $t$, OS concentrates the entire evaluation effort within
$o_{\max}$ of the target size. Both schemes execute every $\ADD{}$/$\RMV{}$
as the exhaustive sweep of Eq.~\eqref{eq:addrmv} --- the economics the
present method changes.

\paragraph{The sequential lineage.}
From individually-best ranking \cite{whitney71,devijver82} through SFS/SBS,
$(l,r)$ and generalized variants, floating search \cite{pudil94}, adaptive
floating \cite{somol99}, and oscillating search (OS; its dynamic variant
DOS) \cite{somol00,somol08dos},
each generation of the family bought
better subsets at higher search cost (Table~\ref{tab:evolution}) --- every
member still sweeps $O(\Dim-d)$ candidates per forward step, which is what
removes the whole family from consideration at $\Dim\gtrsim 10^3$ under wrapper
criteria \cite{kohavi97}. The method proposed here adds the next row of the
table, and the first one in the family's history that \emph{reduces} per-step
cost instead of adding search effort: a budgeted sampled step that breaks
the per-step $O(\Dim)$ sweep and extends the family into dimensionalities
where previously only ranking methods could operate.
\begin{table}[b!]\centering
\caption{Evolution of sequential feature selection, one generation per row.
Every method up to this work evaluates all available candidates at each step;
the proposed budgeted step is the family's first departure from the full
sweep.}\label{tab:evolution}
\begin{tabular}{lll}\toprule
Method & Introduced & What it added \\ \midrule
BIF ranking & \cite{whitney71,devijver82} & individual relevance ordering; no interplay modeled \\
SFS / SBS & \cite{whitney71} & greedy set-level construction (nesting problem) \\
plus-$l$-minus-$r$ & \cite{stearns76} & fixed-depth backtracking against nesting \\
generalized $(l,r)$ & \cite{devijver82} & tuple-wise steps \\
SFFS / SBFS & \cite{pudil94} & self-controlled (floating) backtracking depth \\
ASFFS & \cite{somol99} & adaptive generalization near the target size \\
OS / DOS & \cite{somol00,somol08dos} & oscillation around the target size; $d$-optimization \\
\textbf{budgeted sampled steps} & this paper & \textbf{per-step cost independent of $\Dim$} \\
\bottomrule\end{tabular}\end{table}

\paragraph{Restricting the candidate pool.}
Restricting the candidate pool of a sequential wrapper search to regain
tractability at high dimensionality has an established lineage. Linear Forward
Selection (LFS) \cite{gutlein09lfs} truncates the pool to the top-$k$ features
of a one-shot pre-search ranking (obtained with the same wrapper evaluator,
applied per attribute) and searches \emph{exhaustively} within it; the cap
applies to forward expansions only. LFS is the nearest published point in
method space to ours, and its own analysis anticipates both limitations our
design targets: the authors note that a static cut may lose ``weakly relevant
attributes that perform poorly on their own'' yet help in combination
\cite{gutlein09lfs}, and their remedy --- growing $k$ into the low hundreds ---
works only because the cap cannot adapt. Notably, they also applied their
pool restriction to floating search \cite{pudil94}, but report that it led to
results similar to plain capped forward selection and omit it from the paper
--- consistent with our reading that a \emph{static} pool neutralizes the
floating mechanism (backward excursions cannot recruit outside a fixed pool),
whereas under online-updated sampling the floating mechanism is precisely
where the retained quality comes from (Section~\ref{sec:results}).
Filter-dominating hybrid SFFS \cite{gan14hybridsffs} reportedly gates each
SFFS step's wrapper evaluations by a separability index; hybridization operators
\cite{somol10hybrids,somol06cs} formalize the pattern: a fast auxiliary
criterion pre-ranks all $T$ available candidates, of which only the top
$\lfloor\lambda T\rceil$, $\lambda\in(0,1]$, are re-evaluated by the wrapper. HyCluster
\cite{ganjei22hycluster} confines SFS/IWSS/SFFS to filter-derived clusters;
IGIS \cite{nakariyakul18igis} deterministically re-ranks all candidates by
interaction information at each forward step; PFBP \cite{tsamardinos18pfbp}
prunes candidates by conditional-independence tests in a parallel
forward--backward search. In all of these the restriction is deterministic and
either fixed before the search or recomputed from a statistic independent of
the wrapper's evaluation trajectory; a feature discarded by the gate is, in
most designs, never reconsidered. Our method differs in both respects: the
per-step candidate set is a fresh stochastic sample under a fixed budget
$\budget$, so per-step cost is decoupled from $\Dim$ while every feature
retains nonzero selection probability throughout the search (the uniform
exploration floor), and the proposal distribution is learned \emph{online} from
every wrapper evaluation the search itself performs, so early mis-rankings are
self-correcting rather than permanent. Three further contrasts with
hybridization \cite{somol10hybrids} carry the design: (i) no second criterion
is needed --- the proposal order derives from the main criterion's own history
at zero additional evaluations; (ii) the hybrid wrapper budget
$\lambda T$ remains proportional to the pool while our $\budget$ is constant;
(iii) the hybrid takes a deterministic top-cut where we sample --- our top-$k$
ablation is precisely the hybridization principle with statistics-as-filter,
so the ablation matrix doubles as a head-to-head comparison.

\paragraph{Stochastic-greedy and submodular optimization.}
Replacing the full per-step sweep of greedy selection with a fixed-size random
sample was formalized as \textsc{Stochastic-Greedy} \cite{mirzasoleiman15lazier},
achieving a $(1-1/e-\varepsilon)$ expected approximation for monotone submodular
objectives at runtime independent of the target cardinality; the framework was
extended to weakly submodular statistical objectives \cite{khanna17weaksubmodular}
and to non-monotone, non-submodular objectives under general constraints
\cite{qian18stochasticgreedy}; noise-robust variants exist
\cite{hassidim17noise,fourati23rgl}. These results motivate a fixed per-step
budget, but the transfer to sequential feature selection is not direct:
sampling in this family is strictly uniform, the search is forward-only, and
in-context FS criteria provide no general support for the structural
assumptions --- wrapper accuracy at $\Dim\ge 5{,}000$ satisfies none of them,
and even a monotone filter criterion carries no submodularity guarantee.
Hashemi et al.~\cite{hashemi19tradeoffs} show that uniform sampling at fixed
size fails with high probability to recover the optimal subset unless the
sample is progressively enlarged. Our response to this failure mode is
orthogonal: rather than growing the budget, we concentrate a \emph{fixed}
budget adaptively by learning the proposal distribution online, and we embed
the budgeted step in a floating search whose backward sweeps provide the
subset-repair capability that forward-only stochastic greedy lacks. The
uniform-sampling ablation in Section~\ref{sec:results} is exactly
\textsc{Stochastic-Greedy} transferred to wrapper FS --- and its collapse on
synergistic data is the empirical cost of feature-blind sampling.

\paragraph{Bandit- and RL-guided feature selection.}
Online per-feature statistics guiding candidate choice originate with FUSE
\cite{gaudel10fuse}, which casts FS as a one-player game solved by UCT: RAVE
scores are updated from every rollout and blended with UCB to select the next
feature, with progressive widening bounding the arms considered per node.
Later bandit formulations replace sequential search altogether: combinatorial-bandit
FS \cite{liu21cmabfs}, Thompson-sampling FS \cite{durand14thompson,liu23tvs},
and multi-agent RL \cite{liu19marlfs,liu23marlfstkde} draw candidate subsets
per round from reward- or posterior-updated per-feature distributions. Our
method adopts the same epistemic move --- accumulate cross-evaluation
statistics online, exploit them to focus a stochastic search --- but in a
different algorithmic container and with a different statistic: FUSE's action
space admits only feature additions (no floating exclusion is possible) and
progressive widening grows the candidate set with visit count rather than
fixing it, while the bandit methods have neither sequential structure nor a
per-step evaluation budget. The combinatorial-bandit design \cite{liu21cmabfs}
is illustrative of the difference in granularity: whole subsets of a fixed,
pre-specified cardinality $K$ are drawn each round and evaluated once, and the
per-feature statistics are updated only by whether the round improved on its
predecessor, with credit shared equally across the two consecutive subsets ---
a binary, trial-level signal. The budgeted step instead extracts a graded
signal from \emph{every} candidate evaluation (its batch-standardized
criterion value, credited exactly to the features whose presence or absence
distinguishes the candidate), constructs subsets incrementally across all
cardinalities at once, and inherits the floating repair mechanism --- while
per-step cost stays fixed. We retain the classical floating-search scaffold,
evaluate a fixed-size softmax-sampled batch per forward step against the
search criterion itself (never a surrogate), and drive the sampler with an
online analogue of
dependency-aware feature ranking \cite{somol11daf}, of which this work is a
direct continuation.

\section{Stochastic Sequential Search}
\label{sec:method}
\subsection{Online dependency-aware statistics}
\label{sec:stats}
The proposal distribution is driven by two accumulators per feature:
$\muis{f}$ estimates the mean \emph{standardized} criterion value of the
evaluated subsets that contained $f$, and $\muisnot{f}$ of those that did
not; a feature's sampling score is their contrast,
\begin{equation}
\score{f} = \muis{f}-\muisnot{f},
\end{equation}
the very contrast that dependency-aware ranking (DAF$_0$
\cite{somol11daf}) computes from a dedicated campaign of random probes ---
here accumulated online, from the evaluations the search performs anyway.

The accumulators are updated \emph{batchwise}, not per evaluation. All
criterion values evaluated within one step (and, before the search starts,
within the warm-up round) are buffered; when the step completes, the
buffered batch is z-standardized --- each raw value is replaced by
$z=(\Crit(\cdot)-\hat\mu)/\hat\sigma$, with $\hat\mu,\hat\sigma$ the mean
and standard deviation of that batch --- and only then folded into the
accumulators: for each evaluated subset with standardized score $z$,
\begin{equation}\label{eq:update}
\muis{f} \leftarrow (1-\eta)\,\muis{f} + \eta\, z \quad (f\in\Sub{}),\qquad
\muisnot{f} \leftarrow (1-\eta)\,\muisnot{f} + \eta\, z \quad (f\notin\Sub{}),
\end{equation}
where every accumulator carries its own update count $i$ and rate
$\eta=1/\min(i,h)$: an exact running mean until that accumulator has seen
$h$ updates, an exponentially forgetting average beyond (statistics stay
fresh as the trajectory drifts across cardinalities). Batchwise
standardization is what makes the contrast meaningful: one batch holds
evaluations of a single cardinality in a single search context, so
z-scores are comparable within it, and expressing every update on this
common scale lets $\score{f}$ aggregate evidence across subset sizes
without per-size bins --- raw criterion values, whose scale and level
shift with cardinality, would not. A useful corollary: a feature not yet
observed on one side has that accumulator at $0$, i.e.\ at the batch
average, so unobserved features start from a neutral prior and no
cold-start correction is needed.

\subsection{The budgeted sampled step operator}
\label{sec:operator}
$\sADD{\budget}{c}$ replaces the exhaustive sweep of Eq.~\eqref{eq:addrmv}
by the evaluation of exactly $\budget$ sampled candidates ($c{=}1$
throughout, as before; $\budget$ is the per-step budget, subscripted
$\budget_f$/$\budget_b$ where the two directions must be distinguished;
a candidate pool not exceeding $\budget$ falls back
to the exhaustive sweep, whose evaluations feed the statistics
identically). Applied to $\Sub{d}$, with free pool
$F=\FSet\setminus\Sub{d}$, one step proceeds in four stages:
\begin{enumerate}
\item \emph{exploration floor:} set $u=\lfloor\rho_u\budget+\tfrac12\rfloor$
      and draw $u$ candidates uniformly without replacement from the $4u$
      \emph{least-evaluated} features of $F$ (or all of $F$ if smaller) ---
      those contained in the fewest evaluated subsets so far. The floor is
      defined by these evaluation counts alone; scores play no role in it.
      The floor thus cycles through the least-observed features --- for a
      feature the search has never picked up, the count is simply the
      number of times it has been proposed --- and guarantees every
      feature periodic re-trial no matter how poorly it is currently
      scored; this, not the softmax tail, is what makes early mis-rankings
      self-correcting;
\item \emph{exploitation draw:} draw the remaining $\budget-u$ candidates
      from the rest of $F$ without replacement, with probability
      proportional to $\exp(\score{f}/\tau)$. The temperature $\tau$ is
      re-derived at every step as a robust scale of the remaining pool's
      current scores, $\tau=\mathrm{IQR}/1.349$ (the normal-consistent
      interquartile-range scale estimate, floored so that a constant-score
      pool degrades to uniform draws) --- the sharpness of exploitation is
      thus self-tuning, with no criterion-scale knob to set;
\item \emph{evaluation:} evaluate the $\budget$ candidate subsets
      $\Sub{d}\cup\{f\}$, buffering every criterion value; the
      best-scoring candidate feature is added;
\item \emph{statistics update:} flush the buffered batch through the
      z-standardized update of Section~\ref{sec:stats}.
\end{enumerate}
All random draws of a step happen in stages 1--2, before the first
evaluation, and the scores $\score{f}$ change in stage 4 only --- once per
step, after the last evaluation. Within a step the proposal distribution
is frozen, and a step's proposals depend on the statistics only through
the preceding steps and the warm-up. This timing is forced by the
standardization itself (the batch mean and deviation that define $z$ exist
only once the batch is complete), and it pins the operator's cost at
exactly $\budget$ criterion evaluations per step. Softmax and
z-standardization thus act at opposite ends of the step, on different
objects: the softmax (stage~2) turns accumulated scores into a proposal
distribution, the z-standardization (stage~4) normalizes the step's raw
criterion values before they enter those scores; the two meet only through
$\score{f}$.

The removal operator mirrors the construction.
$\mathrm{sRMV}^{(\budget_b)}_{c}$ applied to $\Sub{d}$ proposes $\budget_b$
removal candidates from within $\Sub{d}$: the floor fraction $\rho_u$
drawn uniformly from the members whose \emph{absence} has been observed
least often (for removal, what happens \emph{without} $f$ is the scarce
information), the remainder with probability proportional to
$\exp(-\score{f}/\tau)$ --- low-scoring members are the preferred removal
candidates --- with $\tau$ re-derived from the member scores as above; the
batch of $\budget_b$ evaluations of $\Sub{d}\setminus\{f\}$ then updates
the statistics exactly as in the forward case. With the defaults used
throughout this paper ($\budget=100$, $\budget_b=50$, $\rho_u=0.2$;
Section~\ref{sec:setup}), a forward step therefore evaluates exactly
$100$ candidates --- $20$ proposed by the floor, $80$ by the softmax ---
and a backward step $50$ ($10+40$). A budgeted variant of \emph{any}
sequential method follows by operator substitution --- the lowercase-s
naming rule of Section~\ref{sec:intro} in formula form:
\begin{multline}\label{eq:family}
\text{s}X \;=\; X \text{ with }
\ADD{c}:=\sADD{\budget_f}{c},\;
\RMV{c}:=\mathrm{sRMV}^{(\budget_b)}_{c}\\
(X \in \{\text{SFS, SBS, SFFS, SBFS, OS, DOS}, \dots\}).
\end{multline}

\subsection{The stochastic instances in full: sSFFS and sOS}
\label{sec:instances}
The substitution rule of Eq.~\eqref{eq:family} is compact but abstract, and
the choice of the host method carries practical weight. We therefore spell
out the two instances a practitioner is most likely to reach for --- in
enough detail to be implemented from this paper alone --- and state why
these two.

\paragraph{sSFFS.}
Substituting the budgeted operators into the floating scheme of
Section~\ref{sec:related} gives, for a target size $t$ and frontier
restriction $\Delta$:
\begin{enumerate}
\item warm-up: evaluate $m_0$ subsets of cardinality $r$ drawn uniformly at
      random; their evaluations, flushed as one batch, seed the statistics
      of Section~\ref{sec:stats} through the same batch-standardized update
      (no other use is made of them);
\item start with $\Sub{c}=\sADD{\budget_f}{c}(\emptyset)$, $d=c$;
\item forward: $\Sub{d+c}=\sADD{\budget_f}{c}(\Sub{d})$, $d\leftarrow d+c$;
\item retreat: repeat $\Sub{d-c}=\mathrm{sRMV}^{(\budget_b)}_{c}(\Sub{d})$,
      $d\leftarrow d-c$, but only while each removal improves the best value
      recorded so far at the lower cardinality;
\item if $d<t+\Delta$ go to step~3; else report the best recorded subset of
      every cardinality reached.
\end{enumerate}
Steps 2--5 restate the classic schedule of Section~\ref{sec:related}
verbatim, and deliberately so: the substitution changes what one step
\emph{costs}, never what the schedule \emph{decides}, so retention against
full SFFS is directly measurable step for step
(Section~\ref{sec:results}). Spelled out at the operator level, one pass
through the loop reads: propose $\budget_f$ additions ($20$ floor $+$
$80$ softmax at the defaults), evaluate them, add the winner, fold the
batch into the statistics; then propose $\budget_b$ removals, evaluate
them, and if removing the winner improves on the best value previously
recorded at the size below, accept it and continue retreating on the same
terms; when the retreat stops, resume forward proposals. Every evaluation
along the way --- forward or backward, winner or not --- passes through
the statistics update, so the proposal distribution keeps learning from
both directions of the search. SFFS is the natural primary host: it is the
quality reference of the sequential family (Section~\ref{sec:intro}), and
we run it forward for two reasons. The first is economic and was stated in
the introduction: sought subsets satisfy $d\ll\Dim$, which forward
construction reaches in $d$ steps at the cheap low cardinalities. The
second is statistical: a forward run keeps every criterion evaluation at
low cardinality, where in-context criterion estimates are most reliable,
whereas a backward-primary run spends its longest phase near full
dimensionality, where small-sample estimation error is at its worst
\cite{raudys06}. Equally important is what a floating run \emph{returns}:
the best subset recorded at every cardinality $1,\dots,t{+}\Delta$, from a
single run; sSFFS inherits this property unchanged, so beyond producing one
subset it traces the whole criterion-vs-size curve --- and where the
criterion admits an optimal dimensionality, the decline of the best-per-size
values past it is visible in the very run that produced the subsets, making
sSFFS a subset-size \emph{exploration} tool (on the criterion axis; the
validated counterpart follows the two-axis discipline of
Section~\ref{sec:setup}).

\paragraph{sOS.}
When the user has already decided which subset size to optimize for, the
oscillating scheme is the practical host: it spends no evaluations
constructing the full trajectory below $t$. Its stochastic counterpart sOS
(named by the rule of Section~\ref{sec:intro}) substitutes the budgeted
operators into the swings of OS (Section~\ref{sec:related}); for a target
size $t$ and a swing-depth limit $o_{\max}$:
\begin{enumerate}
\item warm-up: as in step~1 of the sSFFS scheme above;
\item initialize the \emph{pivot}: evaluate a starting subset of size $t$
      --- classically a ranking prefix; at extreme $\Dim$ the warm-up
      statistics themselves offer a $\Dim$-independent alternative, the
      top-$t$ features by $\score{f}$ --- and set the swing depth $o=1$;
\item down-swing: apply $\mathrm{sRMV}^{(\budget_b)}_{c}$ $o$ times, then
      $\sADD{\budget_f}{c}$ $o$ times, returning to size $t$; if the
      returned subset improves on the pivot, adopt it as the new pivot,
      reset $o=1$, and repeat step~3;
\item up-swing: apply $\sADD{\budget_f}{c}$ $o$ times, then
      $\mathrm{sRMV}^{(\budget_b)}_{c}$ $o$ times; on improvement adopt
      the new pivot, reset $o=1$, and go to step~3;
\item after a failed down/up pair deepen the swings, $o\leftarrow o{+}1$:
      if $o\le o_{\max}$ go to step~3, else report the pivot.
\end{enumerate}
Concretely, at depth $o=1$ a down-swing reads: propose $\budget_b$
removals from the pivot, evaluate them, remove the winner (size $t{-}1$);
propose $\budget_f$ additions, evaluate them, add the winner (back at
size $t$); compare the resulting subset with the pivot --- each of the two
batches updating the statistics as it completes. As in classic OS, only a
completed swing --- back at size $t$ --- is ever compared with the pivot,
never an intermediate subset. A failed pair at
depth $o$ costs at most $o(\budget_f+\budget_b)$ evaluations per direction
--- independent of $\Dim$ --- and every evaluation inside the swings feeds
the same online statistics, so successive swings around the pivot grow
better informed. The experiments of this
paper exercise sSFFS, whose all-sizes output subsumes the fixed-$t$ task;
sOS inherits the operators and statistics unchanged and is available in the
same implementation.

\section{Experimental Setup}
\label{sec:setup}

\paragraph{Datasets.}
Three public datasets cover the three dimensionality regimes the paper argues
about, and deliberately reuse the test bed of the DAF study \cite{somol11daf}
so that head-to-head numbers are directly comparable. \emph{Madelon}
(NIPS-2003 feature selection challenge \cite{guyon05challenge}; $\Dim=500$,
$2{,}000$ samples, two classes) is the retention-and-ablation ground: full
sequential search is still tractable there, so the budgeted search can be
measured against the method it approximates. Its engineered structure --- the
informative features are useful only in combination, hidden among 480
pure-noise probes --- makes it the canonical stress test for
interaction-blind selection. \emph{Gisette} (same challenge; $\Dim=5{,}000$,
$1{,}000$ samples, the highly confusable handwritten digits 4 vs.\ 9) carries
the headline comparison: the target subset sizes (100--300) lie far beyond
the full-SFFS feasibility frontier (Section~\ref{sec:intro}), so ranking
methods are the only competitors that can enter.
\emph{Reuters-21578, Distribution 1.0}\footnote{We acknowledge the use of
the Reuters-21578 text categorization test collection, compiled by David
D.\ Lewis and distributed free for research purposes; the collection is
available at
\url{https://www.daviddlewis.com/resources/testcollections/reuters21578/}.}
(Apt\'e split; $\Dim=10{,}105$ term features, $8{,}941$ documents, 33
categories) is the extreme-dimensionality showcase at $\Dim>10^4$.

\paragraph{Selection and validation protocol.}
The protocol follows \cite{somol11daf}. Each trial draws a random
class-stratified 50/50 train/test split. The selection criterion on madelon
and gisette is the wrapper accuracy of a 1-NN classifier (Euclidean metric,
features scaled to $[0,1]$), estimated by 3-fold cross-validation on the
training half; the held-out half enters exactly once, to validate reported
subsets with the same classifier trained on the full training half. On
reuters, where wrapper evaluation cost at $\Dim>10^4$ would dominate any
search economics, the criterion is the multinomial Bhattacharyya distance,
as in \cite{somol11daf}. All searches run in $d$-optimizing (anytime) mode:
a single run reports the best subset found at every size, harvested from the
anytime log, and full-range runs are terminated once the search frontier
passes the last target size (165 on madelon, 320 on gisette). Every
experiment is repeated over random seeds --- ten on madelon, five on the
gisette campaign, three per cell of the budget-scaling grids --- and one
seed determines both the split and every stochastic draw of the run. Seed 1
re-uses the split of the full-search reference, so seed-1 comparisons are
strict same-split comparisons (the \emph{shared-split} seed); every other
seed draws its own split, and
cross-method comparisons on them are paired per seed. We report
mean$\pm$std over seeds throughout.

\paragraph{Budgeted-search configuration.}
All experiments run the forward-primary variants. The choice is economic,
not methodological: in the very-high-dimensional regime the sought subsets
satisfy $d\ll\Dim$, which forward construction reaches in $d$ steps while
backward-primary construction needs $\Dim-d$; and since per-evaluation cost
grows with subset cardinality, a backward-primary run spends those
$\Theta(\Dim)$ steps at the most expensive cardinalities. The operators
themselves are direction-symmetric (Section~\ref{sec:method}).
A single default configuration, fixed on madelon probes before any gisette
experiment ran (Section~\ref{sec:discussion}), is used for every budgeted
run in the paper: forward budget $\budget=100$, backward budget
$\budget_b=50$, exploration floor $\rho_u=0.2$, automatic robust
temperature, forgetting horizon $h=100$ evaluations, batch-standardized
contrast statistic, and a warm-up of $m_0=200$ probes of cardinality
$r=10$ on madelon --- written $m_0@r$: $200@10$ --- with $1{,}000@25$ on
gisette and $2{,}000@25$ on reuters, the only setting scaled with $\Dim$. Warm-up probes are ordinary criterion
evaluations and are included in every budget count. On madelon all
campaigns additionally run a twin arm with unrestricted backward sweeps
($\budget_b=|\Sub{}|$, the exact floating correction), isolating the cost
and quality effect of budgeting the backward direction.

\paragraph{Baselines and ablation arms.}
Full SFFS and straight SFS provide the sequential quality references where
they are computable (the full madelon size range; on gisette full SFFS
serves only to establish the feasibility frontier). The ranking competitors
are BIF --- exactly $\Dim$ singleton evaluations, a single point on any
budget axis --- and DAF \cite{somol11daf} in its probe-based protocol:
uniformly drawn random subsets of cardinality 1--50 scored by the same
criterion, features ranked by the DAF$_0$ contrast (the difference between
the mean criterion value of probes containing the feature and of those
lacking it), and the ranking's prefixes validated on the held-out half (every size up to 150 on madelon,
every fifth up to 300 on gisette). DAF's consumed budget is exactly its
probe count, which we match per seed to the budgeted search on both
fairness axes: matched evaluation counts and matched wall time. The two
axes differ materially --- probes evaluate small subsets and are far cheaper
per evaluation than the large-subset candidates of a sequential search ---
so both are reported. The ablation arms of Section~\ref{sec:related} each
replace exactly one design ingredient at an otherwise identical
configuration and budget: uniform sampling (\textsc{Stochastic-Greedy}
transferred to wrapper FS), deterministic top-$k$ truncation of the
ranking at $k=\budget$ (the hybridization principle), statistics frozen after warm-up (the
static-ranking principle of LFS), the raw running-mean statistic in place
of the batch-standardized contrast, and a constant-fraction budget schedule
in place of constant count at matched total budget.

\paragraph{Compute accounting and campaign grids.}
Search cost is reported on two axes: criterion-evaluation counts, which are
machine-independent, and cost-weighted time under the measured
single-thread cost model of the reference machine (for gisette,
$\mathrm{cost}(d)\approx 0.79+0.20\,d$~ms per evaluation of a size-$d$
candidate, linear to at least $d=200$). The trade-off campaign on madelon
sweeps $\budget\in\{10,25,50,100,200\}$ under the softmax sampler and
$\budget\in\{25,100\}$ under the uniform and top-$k$ samplers, ten seeds
per cell; the budget-scaling study on gisette runs DAF at probe budgets
$10^3$--$10^6$ against the budgeted search at
$\budget\in\{10,25,50,100,200\}$ (three seeds per cell, five at
$\budget=100$), with BIF's fixed $\Dim$ evaluations as the single ranking
point. The extreme-dimensionality grid on reuters-apte ($10{,}105$ term
features, $8{,}941$ documents, 33 classes, the corpus of \cite{somol11daf};
a single random 50\%/40\% train/test split per seed --- the
random-resampling protocol RR(1,50,40) --- five seeds) switches to the
multinomial Bhattacharyya filter criterion with multinomial-naive-Bayes
(MultinomNB) holdout validation: sSFFS at $\budget\in\{25,100\}$ ($\budget_b=50$, warm
start $2{,}000@25$, frontier-killed at $d\ge500$), the toolbox-native BIF
prefix curve at $d\in\{10,\dots,500\}$, the DAF probe ladder at
$10^4$--$10^6$ trials, and matched-evaluation DAF controls at the sampled
arms' exact totals; measured filter cost
$\approx0.006+0.0005\,d$~ms per evaluation.

\paragraph{Stability and cross-classifier validation.}
Following the protocol of \cite{somol11daf}, accuracy results are
complemented by selection-stability measures between repeated selections ---
the Average Tanimoto Index (ATI), the relative weighted consistency
(CW$_{\mathrm{rel}}$), and the Average Normalized Hamming Index (ANHI)
\cite{somol10pami} --- and by cross-classifier
panels: subsets selected with the 1-NN wrapper are validated by an
RBF-kernel SVM and vice versa, controlling for criterion over-fitting.
Because the warm-start statistics must not leak across repeated splits,
stability is computed from $N$ fully independent runs (one seed = one split
\emph{and} one sampling stream each, warm-up included) rather than from a
shared-state repeat-over-splits protocol; the similarity measures are
evaluated externally by an implementation verified to reproduce the
toolbox's native tracker values exactly on a warm-up-free scenario pair
before use.

\paragraph{Implementation and reproducibility.}
All methods --- the budgeted operators, the sequential and ranking
baselines, and the validation machinery --- are implemented in the
authors' Feature Selection Toolbox 4 (FST4)\footnote{The Feature
Selection Toolbox line resides at \url{http://fst.utia.cas.cz}; FST4
will be published there alongside its predecessors if it is not
available yet at the time of reading.},
and every experiment in this paper is a single invocation of its
command-line tool, reproducible from the
configuration files and command lines accompanying the paper. In addition,
the paper is accompanied by a standalone, dependency-free single-file
implementation of sSFFS covering both experiment criteria (the 1-NN
wrapper with cross-validation and the multinomial Bhattacharyya filter,
including data splitting and the portable random generator), verified to
reproduce the toolbox implementation bit for bit at matched seeds --- the
same data split, search trajectory, selected subsets, criterion values and
evaluation counts --- so the method and its experiments remain
reproducible from the accompanying material alone. Runs are
single-threaded (concurrency only across independent runs), keeping
evaluation counts and wall times cleanly attributable; the machine of
record for all reported timings is a single Apple M1 Max core. The toolbox
carries its own platform-independent pseudo-random generator, so a seed
pins the data split and the entire search trajectory: same-seed runs
reproduce to output precision, verified across operating systems and CPU
architectures on a reference run. The standalone implementation, its
verification harness, and the configuration files and command lines of the
reported experiments accompany this paper as supplementary material.

\section{Results}
\label{sec:results}

Results are reported in three regimes. On madelon ($\Dim=500$), where full
floating search is still computable, the budgeted search is measured against
the method it approximates; on gisette ($\Dim=5{,}000$), where the target
sizes lie beyond the full-search frontier, it is measured against ranking
methods at matched budgets; reuters probes the $\Dim>10^4$ regime.

\begin{figure}[t]
\centering
\begin{tikzpicture}
\begin{axis}[
  width=\linewidth, height=6cm,
  xlabel={subset size $d$}, ylabel={criterion value $\Crit$},
  xmin=1, xmax=165, ymin=0.55,
  legend pos=south east, legend cell align=left,
  every axis plot/.append style={thick, mark=none},
]
\addplot[name path=lo, draw=none, forget plot] table[x=d, y=lo] {figures/data/madelon_g1_per_size_all.tsv};
\addplot[name path=hi, draw=none, forget plot] table[x=d, y=hi] {figures/data/madelon_g1_per_size_all.tsv};
\addplot[blue!20, forget plot] fill between[of=lo and hi];
\addplot+[blue!60, dashed]  table[x=d, y=mean]      {figures/data/madelon_g1_per_size_all.tsv};
\addplot+[black]            table[x=d, y=full_sffs] {figures/data/madelon_g1_per_size_all.tsv};
\addplot+[blue]             table[x=d, y=seed1]     {figures/data/madelon_g1_per_size_all.tsv};
\legend{sSFFS 10-seed mean ($\pm$std band), full SFFS (baseline split), sSFFS seed 1 (same split)}
\end{axis}
\end{tikzpicture}
\caption{Anytime best-per-size criterion value on madelon
(\emph{train-criterion axis}: 1-NN-wrapper 3-fold-CV value on the training
half --- the optimization target): sSFFS
($\budget=100$, $\rho_u=0.2$, warm start $200@10$) retains
$\ge 98.2\%$ of the full-SFFS value at every size (seed 1, identical data split)
and exceeds it at every size above $d=130$; the shaded band is the
mean$\pm$std over 10 seeds (seeds $\ne 1$ on independent splits).}
\label{fig:retention}
\end{figure}
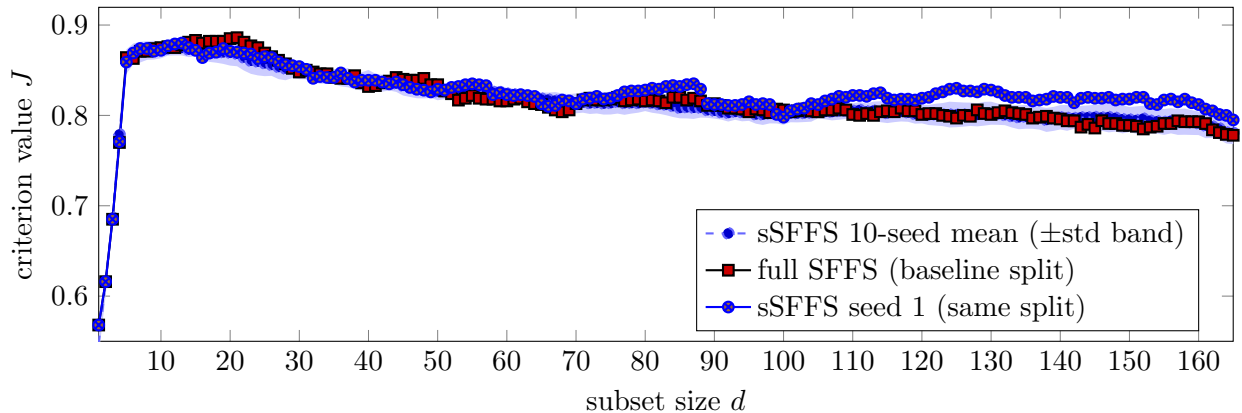

\paragraph{Retention of full-floating-search quality (madelon).}
Figure~\ref{fig:retention} compares sSFFS (forward budget
$\budget=100$; backward sweeps unrestricted in this arm) with the full-SFFS
anytime reference over all subset sizes $1\dots165$. On the shared-split
seed the budgeted search retains at least $98.2\%$ of the full-search
criterion value at \emph{every} size (per-size mean $101.2\%$) --- and above
$d=130$ it exceeds the full search at all 36 sizes, by up to $+0.035$
(Section~\ref{sec:discussion} returns to these greedy-trap escapes). Across
all ten seeds the per-seed mean retention spans $99.0$--$101.2\%$, between
145 and 165 of the 165 sizes lie above the $97\%$ mark per seed, and no
single size falls below $93.5\%$ anywhere in the campaign; the budgeted
searches peak at criterion values $0.866$--$0.892$ at sizes $8$--$20$,
bracketing the full-search optimum ($0.886$ at $d=21$). Seeds $2$--$10$
draw their own data splits, so the band of Figure~\ref{fig:retention} mixes
split variance with sampling variance; the strict like-for-like comparison
is the shared-split seed, and it is also the strongest.

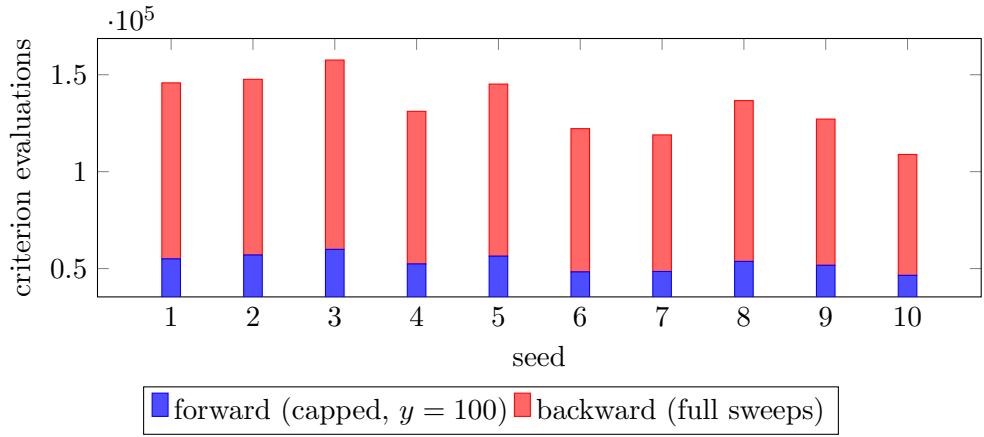
\begin{figure}[t]
\centering
\begin{tikzpicture}
\begin{axis}[
  width=0.8\linewidth, height=5cm,
  ybar stacked, bar width=7pt,
  xlabel={seed}, ylabel={criterion evaluations},
  xtick=data,
  legend to name=leg:anatomy, legend columns=2, legend cell align=left,
]
\addplot+[fill=blue!70]  table[x=seed, y=fwd_evals] {figures/data/madelon_g1_eval_anatomy_all.tsv};
\addplot+[fill=red!60]   table[x=seed, y=bwd_evals] {figures/data/madelon_g1_eval_anatomy_all.tsv};
\legend{{forward (capped, $\budget=100$)}, {backward (full sweeps)}}
\end{axis}
\end{tikzpicture}

\smallskip
\ref{leg:anatomy}
\caption{Where the budget goes under forward-only capping: with forward steps
capped at $\budget=100$, the uncapped backward sweeps of the floating
correction dominate total cost
(57--62\% across all ten seeds) once the search explores sizes up to $d=165$ — the
measured motivation for budgeting the backward direction as well ($\budget_b$)
and for bounded size ranges in the
extreme-dimensionality regime.}
\label{fig:anatomy}
\end{figure}

\paragraph{Evaluation anatomy: budgeting both directions.}
With forward steps capped and backward sweeps left exact, the floating
correction dominates cost: backward evaluations account for $57$--$62\%$ of
the total on every seed (Figure~\ref{fig:anatomy}), at $109$--$158$k
evaluations per seed to cover all sizes up to 165 --- on the shared-split
seed ${\approx}46\%$ of the ${\approx}320$k-evaluation full-SFFS cost model.
Budgeting the backward direction as well ($\budget_b=50$, the default
configuration of Section~\ref{sec:setup}) cuts the totals to $52$--$80\%$
of the forward-only-capped runs (median ${\approx}60\%$; $77.6$k
evaluations on the shared-split seed, ${\approx}24\%$ of the full-SFFS cost
model) while leaving retention essentially unchanged: at least $97.1\%$ at
every size on the shared-split seed (per-size mean $99.9\%$), per-seed mean
retention $98.8$--$100.7\%$ across the ten-seed twin arm. The price is
confined to the high-$d$ stochastic excursion wins: at $d=150$ the
backward-capped run sits at the full-SFFS level ($0.788$ vs.\ $0.789$)
where the uncapped-backward arm had reached $0.818$.

\begin{figure}[t]
\centering
\begin{tikzpicture}
\begin{axis}[
  width=\linewidth, height=5.5cm,
  ybar, bar width=3pt,
  xlabel={seed}, ylabel={criterion value $\Crit$ at $d=20$},
  xtick=data, ymin=0.45, ymax=0.95,
  legend pos=south west, legend cell align=left,
]
\addplot+[fill=black!70]  table[x=seed, y=straight] {figures/data/madelon_d20_collapse_winner.tsv};
\addplot+[fill=blue!70]   table[x=seed, y=sampled]  {figures/data/madelon_d20_collapse_winner.tsv};
\addplot+[fill=red!70]    table[x=seed, y=uniform]  {figures/data/madelon_d20_collapse_winner.tsv};
\legend{full-sweep SFS, softmax sampling (ours), uniform sampling}
\end{axis}
\end{tikzpicture}
\caption{Budget spent uniformly vs.\ informed (madelon, $d=20$, ten paired seeds,
identical budget $\budget=100$ per step; \emph{train-criterion axis} ---
except where a value is explicitly marked as validated holdout accuracy):
softmax sampling matches the full sweep
on every seed (means $0.866$ vs.\ $0.863$), while uniform sampling never wins
against either (vs.\ softmax: nine losses, one tie within $2\cdot10^{-5}$) and
collapses outright on two seeds (5 and 8: criterion $0.62$/$0.75$, validated
accuracy $0.52$/$0.73$) --- madelon's synergistic features are found by the
learned statistics, not by luck. At tighter budgets the uniform failure deepens
to chance level (Discussion).}
\label{fig:collapse}
\end{figure}
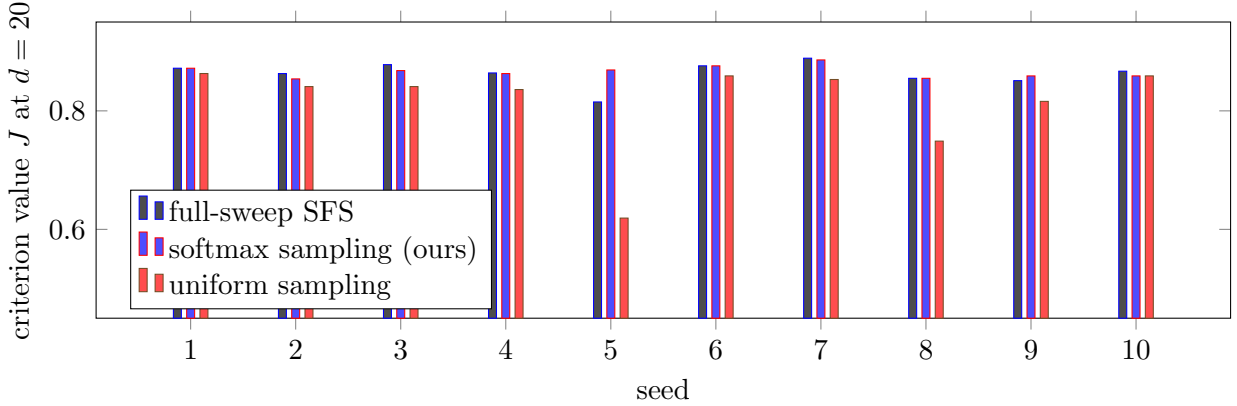

\paragraph{Uniform budget spending collapses; informed spending does not.}
At an identical per-step budget ($\budget=100$, $d=20$ probe, ten paired
seeds), the informed sampler reaches criterion parity with the full-sweep
search (means $0.866$ vs.\ $0.863$; paired $t$-test, $t(9)=0.5$) at $22\%$ of its
evaluations ($2{,}200$ vs.\ $9{,}810$), while uniform sampling at the very
same budget never beats the informed sampler on any seed (nine losses, one
tie; Wilcoxon signed-rank $p<0.01$) and collapses outright on two of ten:
seeds 5 and 8 fall to criterion $0.62$/$0.75$ and validated accuracy
$0.52$/$0.73$, against the informed sampler's $0.85$ on both
(Figure~\ref{fig:collapse}). On madelon's synergistic features an
uninformed budget fails catastrophically, not gracefully, and tightening
the budget deepens the failure ($\budget=50$: uniform mean $0.757$, with
chance-level validated accuracy on one seed) while the informed sampler
degrades gently. The full ablation ladder separating the warm-start ranking
from online adaptation, and the remaining arms (top-$k$ truncation
$0.834$, raw-mean statistic $0.839$, constant-fraction schedule $0.867$),
are analyzed in Section~\ref{sec:discussion}.

\paragraph{Gisette: optimization power at ranking budgets.}
Table~\ref{tab:ladder} reports the budget ladder on both axes of
Section~\ref{sec:setup}: the criterion $\Crit$ (the search objective) and
the holdout accuracy of the same subsets. On the objective, sSFFS
sustains $\Crit\ge 0.98$ at every subset size from ${\sim}50$ to 320 at
\emph{every} per-step budget tried --- including $\budget=10$ --- while the
DAF prefix criterion saturates near $0.94$ (at $10^5$ probes; $10^6$ does
not close the gap) and BIF stalls at $0.90$: the search out-optimizes the
ranking methods by $0.05$--$0.10$ on their own budget scale, at every rung
of the ladder. For perspective, full SFFS would spend ${\sim}175$k
evaluations on forward sweeps alone to reach size ${\sim}35$; the budgeted
runs cover \emph{every} size up to 320 within comparable totals
($83$k--$244$k, warm-up included). A single run-to-completion probe (seed 1,
$\budget=100$) pushes the same anytime frontier on to $d=1{,}000$ within
$491$k evaluations total --- a full sweep spends ${\sim}4.5$M on forward
adds alone to reach that size --- and lifts the best-at-size criterion to
$0.996$ uniformly across $d\in[100,1000]$, in ${\sim}11$~h of wall time on
the reference machine under concurrent campaign load. This is the claim of the paper
delivered on the objective axis: sequential, set-level optimization ---
of whatever criterion the user supplies --- at budgets where previously
only ranking could operate.

\begin{table}[t]\centering
\caption{Gisette budget ladder (mean over seeds; three per cell, five at
$\budget=100$): criterion value $\Crit$ (3-fold CV wrapper on the training
half --- the search objective) and holdout accuracy of the \emph{same}
subsets, at subset sizes 100 and 300. DAF's budget is its probe count; the
sSFFS totals cover the whole anytime run (all sizes $\le 320$,
warm-up included). Reference points: the all-features holdout accuracy is
$0.894$--$0.902$ across the three seeds; matched-budget DAF controls on both
fairness axes --- matched evaluations (138k--233k probes) and matched wall
time ($1.5$--$2.4$M probes), five seeds each --- validate at $0.89$--$0.95$,
on the same saturation plateau as the fixed-budget rows above.}
\label{tab:ladder}
\begin{tabular}{llcccc}\toprule
 & evals & $\Crit$@100 & acc@100 & $\Crit$@300 & acc@300 \\ \midrule
BIF & $5{,}000$ & 0.895 & 0.907 & 0.905 & 0.912 \\ \midrule
DAF $10^3$ & $1{,}000$ & 0.889 & 0.893 & 0.909 & 0.904 \\
DAF $10^4$ & $10{,}000$ & 0.904 & 0.911 & 0.924 & 0.925 \\
DAF $10^5$ & $100{,}000$ & 0.890 & 0.905 & 0.940 & 0.927 \\
DAF $10^6$ & $1{,}000{,}000$ & 0.897 & 0.901 & 0.939 & 0.927 \\ \midrule
sSFFS $\budget=10$ & $196$k & 0.989 & 0.869 & 0.983 & 0.863 \\
sSFFS $\budget=25$ & $83$k & 0.988 & 0.883 & 0.988 & 0.886 \\
sSFFS $\budget=50$ & $111$k & 0.984 & 0.881 & 0.984 & 0.878 \\
sSFFS $\budget=100$ & $176$k & 0.988 & 0.881 & 0.988 & 0.877 \\
sSFFS $\budget=200$ & $244$k & 0.992 & 0.873 & 0.992 & 0.881 \\
\bottomrule\end{tabular}\end{table}

\paragraph{The criterion--generalization gap at $n\ll\Dim$.}
The holdout column of Table~\ref{tab:ladder} inverts the picture, and we
report it as a finding, not a footnote: the subsets that score
$\Crit\approx 0.99$ validate at only $0.86$--$0.89$ --- at or below the
use-all-features reference --- while DAF prefixes validate at
$0.89$--$0.95$ and BIF at $0.91$, both \emph{above} their own criterion
values. With $n=500$ training samples and $\Dim=5{,}000$ candidate features,
an
optimizer this effective finds subsets that exploit the particular sample
--- the harder the criterion is optimized, the larger the share of the
gain that is selection bias --- whereas ranking-by-averaging never
optimizes any individual subset's score and loses nothing on the holdout.
The madelon campaign shows the same machinery with criterion and holdout
in agreement (2{,}000 samples, $\Dim=500$); the gap is a property of the
$n\ll\Dim$ regime and of the criterion, not of the search: full SFFS
would inherit it identically at these sizes but cannot get there to show
it --- the budgeted search is the first member of the family able to
\emph{expose} it at $\Dim=5{,}000$, empirically confirming the
over-fitting concern raised for sequential wrapper search in
\cite{somol11daf}. The practical reading: the method removes the
computational barrier and returns the choice of objective to the
practitioner, for whom criterion design (regularized or filter criteria,
larger validation designs) becomes the binding decision beyond the
frontier. The hypothesized extreme-budget DAF \emph{decline}, finally, is
not supported: per-seed $3{\cdot}10^5\!\to\!10^6$ holdout deltas span
$-0.004$ to $+0.004$, within the resolution of a 500-sample holdout.

\paragraph{Quality vs.\ compute trade-off.}
\begin{figure}[p]
\centering
\pgfmathsetmacro{\costA}{3.6797}
\pgfmathsetmacro{\costB}{0.48717}
\begin{tikzpicture}
\begin{groupplot}[
  group style={group size=2 by 2, horizontal sep=1.2cm, vertical sep=1.35cm},
  width=0.47\linewidth, height=6.2cm,
  xmode=log,
  every axis plot/.append style={thick, mark size=2.4pt},
  legend cell align=left, legend style={font=\footnotesize},
]
\nextgroupplot[title={$d=20$}, ylabel={criterion value $\Crit$},
  xmin=380, xmax=200000, ymin=0.53, ymax=0.91,
  xlabel={evaluations invested (log)},
  legend to name=leg:tradeoff, legend columns=3]
\addplot[gray, thick] coordinates {(380,0.885025) (200000,0.885025)};
\addlegendentry{full SFFS (asymptote)}
\addplot[black, dashed] coordinates {(380,0.798012) (200000,0.798012)};
\addlegendentry{DAF ($10^4$ probes)}
\addplot[black, dotted] coordinates {(380,0.556008) (200000,0.556008)};
\addlegendentry{BIF ($501$ evals)}
\addplot[blue, mark=*, error bars/.cd, y dir=both, y explicit]
  table[x=mean_evals, y=mean_crit, y error=std_crit,
        discard if not={sampler}{softmax}, discard if not={d}{20}]
  {figures/data/tradeoff_slices_agg.tsv};
\addlegendentry{softmax (weighted)}
\addplot[red, mark=square*, error bars/.cd, y dir=both, y explicit]
  table[x=mean_evals, y=mean_crit, y error=std_crit,
        discard if not={sampler}{uniform}, discard if not={d}{20}]
  {figures/data/tradeoff_slices_agg.tsv};
\addlegendentry{uniform}
\addplot[green!50!black, mark=triangle*, error bars/.cd, y dir=both, y explicit]
  table[x=mean_evals, y=mean_crit, y error=std_crit,
        discard if not={sampler}{topk}, discard if not={d}{20}]
  {figures/data/tradeoff_slices_agg.tsv};
\addlegendentry{top-$k$}
\addplot[black, dashed, only marks, mark=diamond*, forget plot]
  coordinates {(10000,0.798012)};
\addplot[black, only marks, mark=pentagon*, forget plot]
  coordinates {(501,0.556008)};
\nextgroupplot[title={$d=100$},
  xmin=380, xmax=200000, ymin=0.53, ymax=0.91,
  xlabel={evaluations invested (log)}]
\addplot[gray, thick, forget plot] coordinates {(380,0.805984) (200000,0.805984)};
\addplot[black, dashed, forget plot] coordinates {(380,0.659019) (200000,0.659019)};
\addplot[black, dotted, forget plot] coordinates {(380,0.550313) (200000,0.550313)};
\addplot[blue, mark=*, forget plot, error bars/.cd, y dir=both, y explicit]
  table[x=mean_evals, y=mean_crit, y error=std_crit,
        discard if not={sampler}{softmax}, discard if not={d}{100}]
  {figures/data/tradeoff_slices_agg.tsv};
\addplot[red, mark=square*, forget plot, error bars/.cd, y dir=both, y explicit]
  table[x=mean_evals, y=mean_crit, y error=std_crit,
        discard if not={sampler}{uniform}, discard if not={d}{100}]
  {figures/data/tradeoff_slices_agg.tsv};
\addplot[green!50!black, mark=triangle*, forget plot, error bars/.cd, y dir=both, y explicit]
  table[x=mean_evals, y=mean_crit, y error=std_crit,
        discard if not={sampler}{topk}, discard if not={d}{100}]
  {figures/data/tradeoff_slices_agg.tsv};
\addplot[black, dashed, only marks, mark=diamond*, forget plot]
  coordinates {(10000,0.659019)};
\addplot[black, only marks, mark=pentagon*, forget plot]
  coordinates {(501,0.550313)};
\nextgroupplot[ylabel={criterion value $\Crit$},
  xmin=0.02, xmax=120, ymin=0.53, ymax=0.91,
  xlabel={cost-model minutes (log)}]
\addplot[gray, thick, forget plot] coordinates {(0.02,0.885025) (120,0.885025)};
\addplot[black, dashed, forget plot] coordinates {(0.02,0.798012) (120,0.798012)};
\addplot[black, dotted, forget plot] coordinates {(0.02,0.556008) (120,0.556008)};
\addplot[blue, mark=*, forget plot]
  table[x expr={(\costA*\thisrow{mean_evals}+\costB*\thisrow{mean_exd})/60000},
        y=mean_crit,
        discard if not={sampler}{softmax}, discard if not={d}{20}]
  {figures/data/tradeoff_slices_agg.tsv};
\addplot[red, mark=square*, forget plot]
  table[x expr={(\costA*\thisrow{mean_evals}+\costB*\thisrow{mean_exd})/60000},
        y=mean_crit,
        discard if not={sampler}{uniform}, discard if not={d}{20}]
  {figures/data/tradeoff_slices_agg.tsv};
\addplot[green!50!black, mark=triangle*, forget plot]
  table[x expr={(\costA*\thisrow{mean_evals}+\costB*\thisrow{mean_exd})/60000},
        y=mean_crit,
        discard if not={sampler}{topk}, discard if not={d}{20}]
  {figures/data/tradeoff_slices_agg.tsv};
\addplot[black, dashed, only marks, mark=diamond*, forget plot]
  coordinates {(2.68,0.798012)};
\addplot[black, only marks, mark=pentagon*, forget plot]
  coordinates {(0.035,0.556008)};
\nextgroupplot[
  xmin=0.02, xmax=120, ymin=0.53, ymax=0.91,
  xlabel={cost-model minutes (log)}]
\addplot[gray, thick, forget plot] coordinates {(0.02,0.805984) (120,0.805984)};
\addplot[black, dashed, forget plot] coordinates {(0.02,0.659019) (120,0.659019)};
\addplot[black, dotted, forget plot] coordinates {(0.02,0.550313) (120,0.550313)};
\addplot[blue, mark=*, forget plot]
  table[x expr={(\costA*\thisrow{mean_evals}+\costB*\thisrow{mean_exd})/60000},
        y=mean_crit,
        discard if not={sampler}{softmax}, discard if not={d}{100}]
  {figures/data/tradeoff_slices_agg.tsv};
\addplot[red, mark=square*, forget plot]
  table[x expr={(\costA*\thisrow{mean_evals}+\costB*\thisrow{mean_exd})/60000},
        y=mean_crit,
        discard if not={sampler}{uniform}, discard if not={d}{100}]
  {figures/data/tradeoff_slices_agg.tsv};
\addplot[green!50!black, mark=triangle*, forget plot]
  table[x expr={(\costA*\thisrow{mean_evals}+\costB*\thisrow{mean_exd})/60000},
        y=mean_crit,
        discard if not={sampler}{topk}, discard if not={d}{100}]
  {figures/data/tradeoff_slices_agg.tsv};
\addplot[black, dashed, only marks, mark=diamond*, forget plot]
  coordinates {(2.68,0.659019)};
\addplot[black, only marks, mark=pentagon*, forget plot]
  coordinates {(0.035,0.550313)};
\end{groupplot}
\end{tikzpicture}

\smallskip
\ref{leg:tradeoff}
\caption{Budget-response of sSFFS on madelon
(\emph{train-criterion axis}: wrapper-1NN 3-fold CV value on the training half
--- the optimization target; validated holdout accuracies are quoted in the
text). Each point: 10-seed mean ($\pm$std in the top row) of the best
criterion value attained at size $d$, versus the compute invested up to that
discovery (warm-up included); the per-step budget grows along each curve
($\budget\in\{10,25,50,100,200\}$ softmax, $\{25,100\}$ controls).
\emph{Top row:} criterion evaluations. \emph{Bottom row:} cost-model minutes
--- evaluations weighted by the measured size-dependent evaluation cost
$\mathrm{ms}(d)\approx3.68+0.487\,d$, which penalizes evaluations spent at
large subset sizes and widens the softmax lead (uniform reaches its
best-at-20 not only later but at larger evaluated sizes). Horizontal
references: full-SFFS anytime value (identical seed-1 split) and the DAF/BIF
rankings at their own budgets (markers). Every sampled arm dominates both
rankings on the criterion at both sizes; softmax reaches its quality with the
least compute throughout and approaches (at $d=100$: matches) the full-sweep
asymptote.}
\label{fig:tradeoff}
\end{figure}
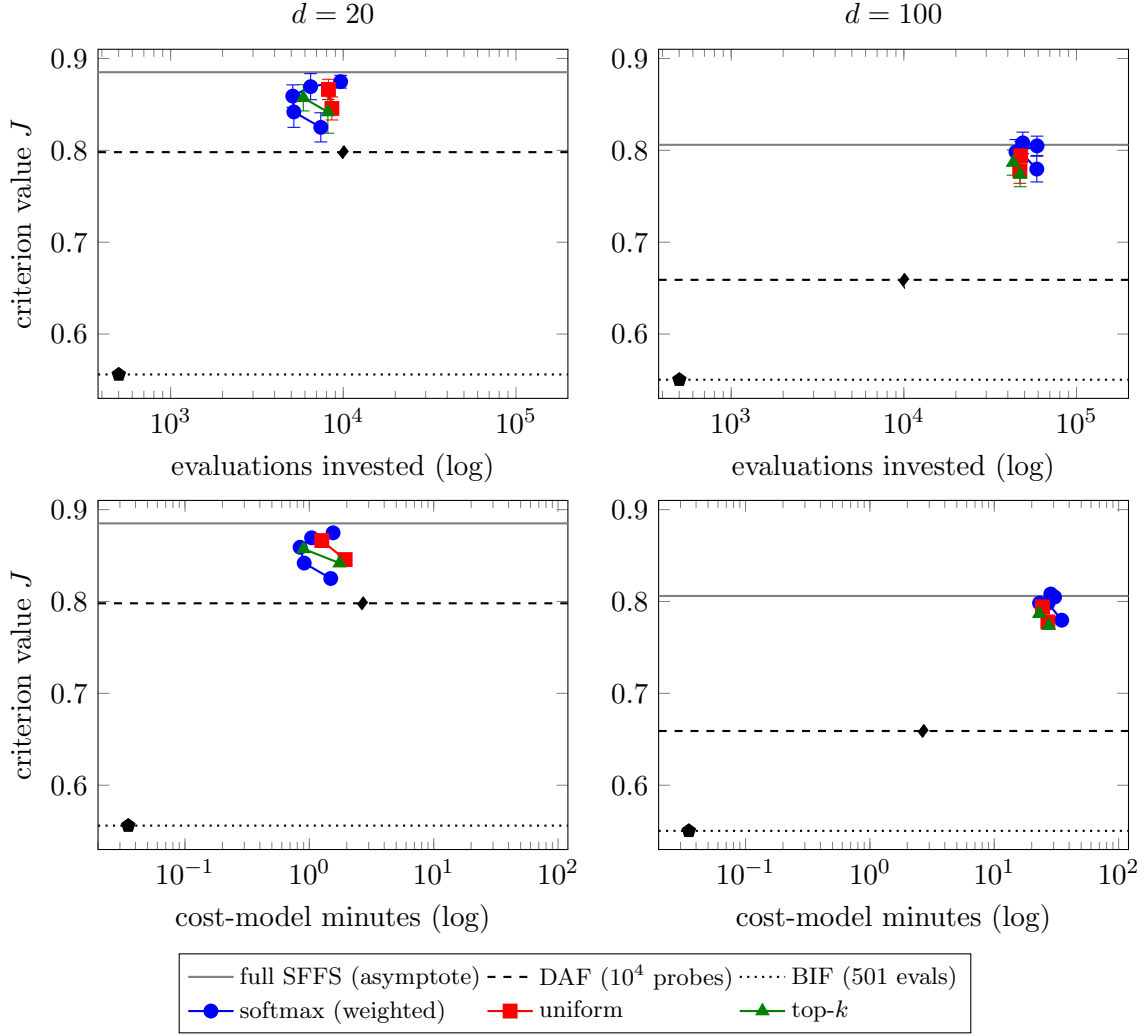

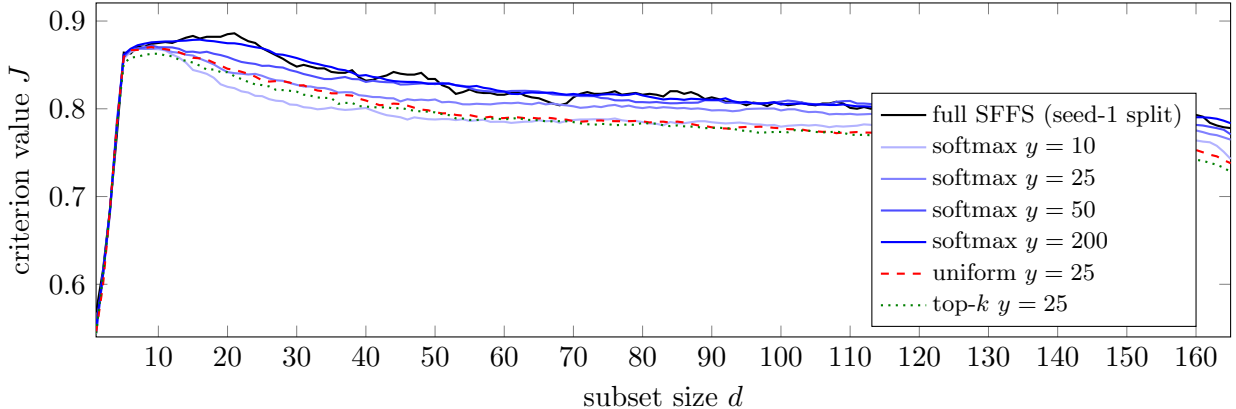
\begin{figure}[t]
\centering
\begin{tikzpicture}
\begin{axis}[
  width=\linewidth, height=6cm,
  xlabel={subset size $d$}, ylabel={criterion value $\Crit$},
  xmin=1, xmax=165, ymin=0.54,
  legend pos=south east, legend cell align=left,
  legend style={font=\footnotesize},
  every axis plot/.append style={thick, mark=none},
]
\addplot[black]         table[x=d, y=full]       {figures/data/tradeoff_persize_wide.tsv};
\addplot[blue!30]       table[x=d, y=softmax10]  {figures/data/tradeoff_persize_wide.tsv};
\addplot[blue!50]       table[x=d, y=softmax25]  {figures/data/tradeoff_persize_wide.tsv};
\addplot[blue!70]       table[x=d, y=softmax50]  {figures/data/tradeoff_persize_wide.tsv};
\addplot[blue]          table[x=d, y=softmax200] {figures/data/tradeoff_persize_wide.tsv};
\addplot[red, dashed]   table[x=d, y=uniform25]  {figures/data/tradeoff_persize_wide.tsv};
\addplot[green!50!black, dotted] table[x=d, y=topk25] {figures/data/tradeoff_persize_wide.tsv};
\legend{full SFFS (seed-1 split), softmax $\budget=10$, softmax $\budget=25$,
        softmax $\budget=50$, softmax $\budget=200$,
        uniform $\budget=25$, top-$k$ $\budget=25$}
\end{axis}
\end{tikzpicture}
\caption{The fixed-$d$ slices of Figure~\ref{fig:tradeoff} generalize to the whole
per-size frontier (madelon, 10-seed mean anytime best-per-size,
\emph{train-criterion axis}): the softmax family closes monotonically towards
the full-SFFS curve as the per-step budget $\budget$ grows (the $\budget=200$
mean overtakes it near the frontier, $d\ge162$), while at the matched tightest
budget the uniform and top-$k$ controls trail softmax $\budget=25$
at $149/165$ resp.\ $164/165$ sizes.}
\label{fig:tradeoff-persize}
\end{figure}

Figures~\ref{fig:tradeoff} and~\ref{fig:tradeoff-persize} map the
budget-response of the method on madelon: ten seeds per cell, SFFS mode,
runs frontier-killed at $d\ge165$, softmax at
$\budget\in\{10,25,50,100,200\}$ against the uniform and top-$k$ controls at
$\budget\in\{25,100\}$, with DAF, BIF and the full-SFFS anytime curve as
references. Four readings. \emph{(i)~Sequential search dominates ranking on
the criterion axis wherever it can reach:} every sampled arm --- down to the
weakest, $\budget=10$ --- exceeds the DAF anchor at both slice sizes, and the
margin at $d=100$ is drastic ($0.77$--$0.81$ vs.\ $0.66$); the weakest arm
passes DAF's $d=20$ value while spending fewer evaluations
(${\sim}7{,}400$) than DAF's own $10^4$ probes. \emph{(ii)~Informed spending
beats blind spending at matched budget:} at $\budget=25$ softmax leads
uniform/top-$k$ at $149/165$ resp.\ $164/165$ subset sizes; at $d=100$ the
value gap is $0.798$ vs.\ $0.778$/$0.774$. At $d=20$, where all samplers
eventually converge in value after a full frontier run, the difference moves
to the compute axis: softmax finds its best-at-20 after ${\sim}5{,}200$
evaluations, uniform needs ${\sim}8{,}600$ --- two-thirds more --- and spends
them at larger subset sizes (eval-weighted mean size $20$ vs.\ $14$), which
the cost-weighted axis penalizes further. \emph{(iii)~The budget knob spans
the gap to the full sweep:} at $d=20$ the softmax mean rises monotonically
from $0.825$ ($\budget=10$) to $0.875$ ($\budget=200$) against the full-SFFS
$0.885$; at $d=100$, $\budget\ge50$ already matches the full sweep ($0.808$
vs.\ $0.806$). \emph{(iv)~On the validated axis} (standing convention: the
figures show the train criterion; holdout follows here) the $d=20$ criterion
gains carry to holdout --- mean 1-NN holdout accuracy rises $0.782\to0.837$
with $\budget$, against DAF $0.820$ (seed-1), BIF $0.561$ (3-seed mean), full
SFFS $0.848$ --- while at $d=100$ every selector from BIF to the full sweep
degrades together (sampled arms $0.62$--$0.65$, full $0.658$, DAF $0.624$,
BIF $0.56$): on madelon's ${\sim}20$ informative features the $d=100$
overfit is driven by subset size, not by the selector.
The cost-weighted view (Figure~\ref{fig:tradeoff}, bottom row) sharpens
rather than softens these readings: weighting each evaluation by the
measured size-dependent cost ($\mathrm{ms}(d)\approx3.68+0.487\,d$ on the
reference machine) penalizes budgets spent at large subset sizes, so the
uniform sampler's disadvantage at $d=20$ roughly doubles (it reaches its
best-at-20 after $2.1\times$ softmax's compute in minutes vs.\ $1.7\times$
in evaluations), while the DAF anchor moves right of every softmax arm's
best-at-20 discovery point.

\paragraph{Extreme dimensionality (reuters).}
\begin{table}[t]\centering
\caption{Reuters ($\Dim=10{,}105$, 33 classes, RR(1,50,40) split): criterion
value $\Crit$ (multinomial Bhattacharyya, computed on the training part ---
the search objective) and MultinomNB holdout accuracy of the \emph{same}
subsets; means over five seeds. Budgets: BIF spends $\Dim$ singleton
evaluations;
DAF its probe count; the sSFFS totals cover the whole anytime run to
the $d_{\max}=500$ frontier (warm-up included). DAF probes here use
cardinality 500 (a CLI constraint pins probe size to the reported subset
size for monotonic criteria); the protocol-conforming card-25 rerun trails
or at best matches card-500 at every ladder budget ($d{=}100$ holdout
$.757/.818/.816$ vs.\ $.801/.831/.837$ at $10^4/10^5/10^6$ probes), so the
table
reports the stronger ranking baseline. ``DAF matched'' re-ranks with card-25
probes at each seed's exact $\budget{=}100$ anytime total
($0.77$--$0.90$M evaluations, mean $0.81$M); the same control at the
$\budget{=}25$ totals ($2.40$--$3.03$M, mean $2.78$M) reproduces the
plateau (see text).}
\label{tab:reuters}
\small
\begin{tabular}{lrrrrrrrrr}\toprule
 & & \multicolumn{2}{c}{$d=25$} & \multicolumn{2}{c}{$d=100$} & \multicolumn{2}{c}{$d=300$} & \multicolumn{2}{c}{$d=500$}\\
 & evals & $\Crit$ & acc & $\Crit$ & acc & $\Crit$ & acc & $\Crit$ & acc\\
\midrule
sSFFS $\budget{=}100$ & $0.81$M & $2.06$ & $.748$ & $3.70$ & $.867$ & $5.17$ & $.907$ & $5.76$ & $.915$\\
sSFFS $\budget{=}25$  & $2.78$M & $2.05$ & $.750$ & $3.70$ & $.867$ & $5.17$ & $.907$ & $5.72$ & $.914$\\
BIF                           & $10{,}105$ & $1.38$ & $.694$ & $3.21$ & $.829$ & $4.79$ & $.886$ & $5.48$ & $.899$\\
DAF matched                   & $0.81$M & $1.29$ & $.647$ & $3.03$ & $.817$ & $4.66$ & $.881$ & $5.36$ & $.898$\\
DAF $10^6$                    & $10^6$  & $1.33$ & $.670$ & $3.29$ & $.837$ & $4.75$ & $.885$ & $5.41$ & $.901$\\
DAF $10^4$                    & $10^4$  & $1.35$ & $.676$ & $2.57$ & $.801$ & $3.23$ & $.843$ & $3.66$ & $.861$\\
\bottomrule
\end{tabular}
\end{table}
The extreme-dimensionality showcase runs the full two-axis protocol on the
reuters-apte corpus ($\Dim=10{,}105$ term features, $8{,}941$ documents, 33
highly imbalanced classes): sSFFS in the winner configuration
($\budget=100$, $\budget_b=50$, warm start $2{,}000@25$) plus a $\budget=25$
arm, frontier-killed at $d\ge500$, against the toolbox-native BIF prefix curve
and the DAF probe ladder, all under the \emph{multinomial Bhattacharyya
filter} with MultinomNB holdout validation --- deliberately a trustworthy
criterion with no cross-validation-proxy overfitting mechanism, to test
whether the optimization advantage transfers to holdout once the criterion
deserves trust. Table~\ref{tab:reuters}: it transfers intact. \emph{The
budgeted search dominates both rankings on both axes at every subset size},
from $d=10$ (holdout $0.630$ vs.\ BIF's $0.543$) through the practically
relevant middle ($d=100$: $+3.8$ points over BIF, $+3.0$ over DAF at
$10^6$ probes) to the frontier ($d=500$: $0.915$ vs.\ $0.899$/$0.901$) ---
the mirror image of the gisette regime, isolating the earlier inversion in
the criterion, exactly where Section~\ref{sec:discussion} places it. Two
further readings. First, ranking saturates: DAF at $10^6$ probes buys
nothing over BIF's $10^4$ evaluations, and the matched-evaluation control
makes the comparison exact --- DAF re-run at each seed's $\budget{=}100$
totals lands on that same plateau (``DAF matched'' row), so at
\emph{identical} spend the sequential search is ahead on both axes at every
size, by $+5.0$ validated points at $d=100$ and $+1.7$ at the frontier.
The control repeats at the $\budget{=}25$ totals ($2.40$--$3.03$M card-25
probes per seed, mean $2.78$M): $3.4\times$ the spend moves the re-ranking
nowhere (holdout $.816$ at $d{=}100$, $.898$ at $d{=}500$), and the
$\budget{=}25$ search at that same spend keeps the two-axis lead
($+5.1$ validated points at $d{=}100$, $+1.6$ at the frontier; criterion
$3.70$ vs.\ $3.02$ and $5.72$ vs.\ $5.40$). Second, on this corpus the budget
knob buys wall-time rather than quality: the $\budget=25$ arm matches
$\budget=100$ at every size but consumes $3.4\times$ the evaluations to
reach the same frontier (backward retreats dominate at ${\sim}3.8$k
evaluations per net feature). The practicality claim is concrete: at the
measured filter cost ($\approx0.006+0.0005\,d$~ms per evaluation) the
entire $\budget=100$ frontier run is roughly two minutes of single-core
evaluation work --- sequential, interaction-aware selection at
$10^4$ dimensions, previously the exclusive regime of rankers, is
routine.

\section{Discussion}
\label{sec:discussion}

\paragraph{Which knobs matter.}
On the madelon $d=20$ probe battery (10 paired seeds), the per-step budget
$\budget$ is the dominant lever, and \emph{allocation
beats volume}: at a matched total budget of ${\sim}2{,}000$--$3{,}000$
evaluations, doubling the per-step cap ($\budget=50\to100$) lifted the mean
criterion value from $0.842$ to $0.866$ --- per-seed parity with the
full-sweep search at ${\sim}22\%$ of its evaluations --- while quintupling
the warm-start probe count on top of that bought nothing measurable
($0.865$). The exploration floor is mildly asymmetric: $\rho_u=0.2$ was never
worse than $0.1$. The robust automatic temperature ($\tau$ from the
interquartile range of current scores) required no tuning in any experiment;
the forgetting horizon default ($h=100$ evaluations) was likewise never the
binding constraint. The full-scale sweep behind
Figure~\ref{fig:tradeoff} confirms both probe-level readings at ten seeds per
cell over the whole $d\le165$ frontier: the budget is the dominant axis and
the sampler the second (softmax $>$ uniform $>$ top-$k$ at matched budget on
$\ge90\%$ of subset sizes, both at $\budget=25$ and $\budget=100$);
$\rho_u$, $\tau$ and $h$ were held at their defaults there, having shown no
sensitivity at probe level.

\paragraph{What carries the quality.}
The ablation ladder decomposes the method's gain over uniform budget spending
into its two ingredients. On the $d=20$ madelon probe (10 paired seeds, all
arms at the winner budget $\budget=100$): uniform sampling $0.814$ $\to$
frozen warm-start-only ranking $0.852$ $\to$ online-updated statistics
$0.866$. The warm-start DAF-style ranking recovers most of the signal
($+0.038$); the online in-search updating adds a further significant
increment ($+0.015$, wins on 9 of 10 seeds, paired $t(9)=3.1$). The gap
widens sharply as the budget tightens: at $\budget=50$ the informed sampler
still holds $0.842$, while uniform sampling falls to $0.757$ with
chance-level validated accuracy on one seed ($0.499$) --- informed sampling
degrades gracefully with budget, uniform sampling catastrophically. Neither
ingredient alone explains the method: the frozen variant
is significantly inferior to online updating, and uniform sampling without
either exhibits the catastrophic collapses of
Section~\ref{sec:results}. Budget schedule (constant count vs.\ constant
fraction of the shrinking pool) proved quality-neutral at matched total
budget on this probe --- but the probe's pool shrinks only $4\%$, so the
comparison cannot strongly discriminate; the constant-count schedule is
preferred on grounds independent of quality: its per-step cost does not scale
with $\Dim$, which is the property that admits extreme-dimensionality
operation at all.

\paragraph{Budgeting the backward direction.}
Measured evaluation anatomy of forward-only budgeting in full-range
floating search is unambiguous: uncapped backward sweeps consume 57--62\% of
all criterion evaluations on every one of ten madelon seeds --- forward-only
capping caps the minority of the cost. Budgeting both directions
($\budget_b=50$) cut total cost to 52--80\% of the forward-only-capped
runs' evaluations to the same
size frontier (median ${\approx}60\%$; 53\% on the shared-split seed,
${\approx}24\%$ of the full-SFFS cost model) while retaining
$\ge 97.1\%$ of the full-SFFS value at every size on the shared-split seed;
the observable price was the loss of some high-$d$ stochastic excursion wins
(at $d=150$: $0.788$ vs.\ the uncapped run's $0.818$, back at the full-SFFS
level). At extreme $\Dim$ the backward share only grows
(a full backward sweep at subset size $d$ costs $d$ evaluations,
${\sim}d_{\max}^2/2$ cumulatively over a frontier reaching size $d_{\max}$),
making direction symmetry a requirement, not an option, for
run-to-completion claims.

\paragraph{Predictable spend.}
A corollary of operator substitution deserves separate statement. Full
SFS/SBS have always had deterministic schedules --- reaching size $d$ costs
exactly $\sum_{k=0}^{d-1}(\Dim-k) = d\Dim-d(d-1)/2$ evaluations, known in
advance --- but the known number is quadratic: $\Theta(\Dim^2)$ whenever the
useful $d$ grows with $\Dim$ or the full size range is swept. At very high
$\Dim$ this is precisely what renders the guarantee empty --- the schedule is
certain and certainly unaffordable. The budgeted operators keep the certainty
and break the quadratic growth: sSFS to size $d$ costs exactly
$\budget_f\,d$ evaluations, independent of $\Dim$ altogether (sSBS
$\budget_b(\Dim-d)$, linear where the sweep was quadratic), and since
per-evaluation cost depends on subset size rather than dimensionality,
completion at any $\Dim$ is guaranteed, not estimated; an evaluation budget
$B$ translates exactly into $\budget_f=B/d$. This is the fixed-spend class in
which only the ranking methods have operated (BIF: exactly $\Dim$
single-feature evaluations; DAF: a user-set probe count), now available with
set-level conditional decisions; of the pool-restriction predecessors only
LFS offers a comparable bound, at the price of the permanent truncation its
authors themselves document \cite{gutlein09lfs}. The floating variants
studied here trade the per-run guarantee away --- backward-correction counts
are data-dependent --- retaining the per-step bound and the anytime mode
instead.

\paragraph{Stochasticity as an occasional asset.}
On individual seeds the budgeted search occasionally \emph{beats} the full
sweep it approximates (up to $+0.035$ at a single size on the shared-split
madelon seed, which exceeds full SFFS at every size from 130 to 165 in the
uncapped-backward runs). These
are greedy-trap escapes --- the sampler declines a locally-best candidate the
deterministic sweep is obliged to take --- and we report them as a curiosity,
not a claim: in expectation, where the full method is tractable, it wins.

\paragraph{Generalization and stability.}
Two diagnostics test whether the budgeted search's stronger optimization
merely over-fits the wrapper criterion. First, \emph{cross-classifier
transfer}: subsets selected with the 1-NN wrapper are validated by an
RBF-kernel SVM and vice versa \cite{somol11daf}, so selection quality is
judged by a classifier that took no part in it. On gisette the 1-NN-selected
$d{=}100$ subsets (train criterion $0.973$) carry a mean $88.7\%$ SVM holdout
accuracy over three independent splits ($90.2/87.8/88.2$) --- the selection
transfers nearly intact to an unrelated model. On madelon the same direction
retains $77.6\%$ (train criterion $0.865$); the reverse direction --- SVM-RBF
selection ($0.665$) validated by 1-NN --- falls to $54.3\%$, near chance, the
honest asymmetry of a synthetic set whose margin-tuned features need not
serve a local classifier. Where transfer is meaningful it survives the change
of classifier.

Second, \emph{selection stability} across independent runs
(ATI/CW$_{\mathrm{rel}}$/ANHI \cite{somol10pami}) marks the one axis on which
the budgeted search does \emph{not} dominate. On madelon ($\Dim{=}500$,
$d{=}20$, ten runs) it is modestly the least stable of the three
(CW$_{\mathrm{rel}}$ $0.654$; ATI $0.490$) against full floating search
($0.760$; $0.617$) and DAF ($0.750$; $0.604$) --- the sampler's stochasticity
paid for directly. On gisette ($\Dim{=}5{,}000$, $d{=}100$) the gap turns
extreme: CW$_{\mathrm{rel}}$ collapses to $0.036$ against DAF's $0.880$, the
budgeted runs sharing almost no features pairwise. This is \emph{solution
multiplicity}, not selection failure: the five runs reach near-identical
criterion values ($0.988\pm0.005$) through nearly disjoint subsets that
nonetheless occupy similar-quality neighbourhoods (ANHI $0.961$, against DAF's
$0.995$) --- the signature of a redundant high-dimensional space holding many
equally-optimal subsets. A stochastic optimizer legitimately samples among
them; a fixed ranking is stable because it is deterministic by construction
\cite{kuncheva07}, not because it has located a uniquely correct subset.
Stability and optimization power are distinct axes, and the budgeted search
trades the first for the second --- most sharply where redundancy renders the
trade harmless.

\paragraph{Limitations.}
The method inherits floating search's lack of optimality guarantees and adds
sampling variance to it; the retention results are empirical, and the
submodular-style guarantee of the uniform special case
\cite{mirzasoleiman15lazier} does not transfer to the adaptive sampler.
Conversely, stronger optimization of an unreliable criterion transfers the
criterion's unreliability more efficiently: on small-sample
high-dimensional data the budgeted search over-fits the wrapper criterion
exactly as full sequential search would (Section~\ref{sec:results}) ---
criterion design remains the practitioner's responsibility, and the
selection-stability protocol above is the corresponding diagnostic.
Warm-start probes are incompatible with protocols that repeat selection over
independent splits within one run (statistics would leak across splits);
per-split re-warming is the clean alternative. Batches must hold candidate
subsets of equal cardinality for the within-batch standardization to be
meaningful --- generalized ($c$-tuple, $c>1$) steps therefore currently fall
back to full sweeps.

\section{Conclusion}
\label{sec:conclusion}

Sequential feature selection has been excluded from very-high-dimensional
problems by the per-step full sweep that every member of the family inherits.
Replacing the sweep with a fixed-size, online-informed stochastic sample ---
dependency-aware statistics collected free of charge from the search's own
evaluations, exploited through a softmax with an exploration floor, in both
search directions --- decouples per-step cost from dimensionality while
retaining, on evidence so far, at least 97\% of full floating-search quality
at a quarter of its evaluations where the full search is still computable.
The uniform and frozen ablations demonstrate that both the informedness and
the continual adaptation of the sampler are necessary: uniform budget
spending collapses on synergistic data and static rankings lose measurably.
Beyond the frontier, on 5{,}000-dimensional gisette, the budgeted search
optimizes the wrapper criterion decisively above the saturation level of
DAF and BIF at matched budgets --- and the accompanying holdout analysis
shows, at a scale no member of the sequential family could previously
reach, that with 500 training samples the criterion itself, not the
search, becomes the binding constraint.
And on $10{,}105$-dimensional reuters, under a filter criterion that
deserves trust, the same search dominates BIF and DAF on the criterion
\emph{and} the holdout at every subset size for about two minutes of
single-core evaluation work. Beyond the frontier where full search cannot enter at all, the
comparison that matters is against ranking methods at matched budgets ---
budgeted sequential search is positioned as the enabler of set-level,
interaction-aware selection in the regime where interaction-blind rankers
were previously the only choice --- matching, in its non-floating form, even
their hard a priori spend guarantees.

\section*{Declaration on the use of generative AI}
During the preparation of this work the authors made substantial use of a
large language model (Claude Fable 5, Anthropic): to draft and revise the
manuscript under the authors' direction, to implement the method and the
experiment tooling (Section~\ref{sec:setup}) and the accompanying
standalone code, and to
orchestrate the experimental
campaigns and harvest their results. The method, the experimental design,
and all scientific claims are the authors' own. Every reported number
originates from deterministic, seed-pinned runs; the harvest tooling was
verified against the toolbox's native output, and the accompanying
standalone implementation reproduces the toolbox bit for bit
(Section~\ref{sec:setup}). The authors reviewed all generated content and
take full responsibility for the content of this publication.

\bibliographystyle{unsrt} %
\bibliography{refs}
\end{document}